\documentclass[runningheads]{llncs}
\usepackage{graphicx}

\usepackage{tikz}
\usepackage{comment} 
\usepackage{amsmath,amssymb} 
\usepackage{color}

\usepackage{graphicx}
\usepackage{xcolor}
\usepackage{algorithm}
\usepackage{algorithmic}
\usepackage{multirow}
\usepackage{array}

\usepackage[colorlinks,linkcolor=red]{hyperref}

\begin{document}
\pagestyle{headings}
\mainmatter
\def\ECCVSubNumber{338}  

\title{Multiple Expert Brainstorming for Domain Adaptive Person Re-identification}

\titlerunning{Multiple Expert Brainstorming for Domain Adaptive Person Re-ID}
%
%
\author{Yunpeng Zhai\inst{1,2} \and
Qixiang Ye\inst{3} \and
Shijian Lu\inst{4} \and
Mengxi Jia\inst{1} \and
Rongrong Ji\inst{5} \and
Yonghong Tian\inst{2}\thanks{Corresponding author.} }

\authorrunning{Y. Zhai et al.}
%
\institute{School of Electronic and Computer Engineering, Peking University, China \and
Department of Computer Science and Technology, Peking University, China \and
University of Chinese Academy of Sciences, China \and
Nanyang Technological University, Singapore \and
Xiamen University, China \\
\email{\{ypzhai, mxjia, yhtian\}@pku.edu.cn, qxye@ucas.ac.cn, \\
shijian.lu@ntu.edu.sg, rrji@xmu.edu.cn
}}

\maketitle

\begin{abstract}

%
Often the best performing deep neural models are ensembles of multiple base-level networks, nevertheless, ensemble learning with respect to domain adaptive person re-ID remains unexplored. 
In this paper, we propose a multiple expert brainstorming network (MEB-Net) for domain adaptive person re-ID, opening up a promising direction about model ensemble problem under unsupervised conditions. 
MEB-Net adopts a mutual learning strategy, where multiple networks with different architectures are pre-trained within a source domain as expert models equipped with specific features and knowledge, while the adaptation is then accomplished through brainstorming (mutual learning) among expert models.
MEB-Net accommodates the heterogeneity of experts learned with different architectures and enhances discrimination capability of the adapted re-ID model, by introducing a regularization scheme about authority of experts. 
Extensive experiments on large-scale datasets (Market-1501 and DukeMTMC-reID) demonstrate the superior performance of MEB-Net over the state-of-the-arts. Code is available at \href{https://github.com/YunpengZhai/MEB-Net} {\color{magenta}https://github.com/YunpengZhai/MEB-Net}.

\end{abstract}

\section{Introduction}
Person re-identification (re-ID) aims to match persons in an image gallery collected from non-overlapping camera networks \cite{yang2019attention}, \cite{jia2020similarity}, \cite{jin2020style}. It has attracted increasing interest from the computer vision community thanks to its wide applications in security and surveillance. Though supervised re-ID methods have achieved very decent results, they often experience catastrophic performance drops while applied to new domains.
Domain adaptive person re-ID that can well generalize across domains remains an open research challenge.

Unsupervised domain adaptation (UDA) in re-ID has been studied extensively in recent years.
Most existing works can be broadly grouped into three categories. The first category attempts to align feature distributions between source and target domains \cite{Wang_2018_CVPR}, \cite{yang2020part}, aiming to minimize the inter-domain gap for optimal adaptation. The second category addresses the domain gap by employing generative adversarial networks (GAN) for converting sample images from a source domain to a target domain while preserving the person identity as much as possible \cite{Liu_2019_CVPR}, \cite{Deng_2018_CVPR}, \cite{Wei_2018_CVPR}, \cite{DBLP:conf/ksem/LvW18}. To leverage the target sample distribution, the third category adopts self-supervised learning and clustering to predict pseudo-labels of target-domain samples iteratively to fine-tune re-ID models \cite{Zhai_2020_CVPR}, \cite{jin2020global}, \cite{DBLP:journals/corr/FanZY17}, \cite{DBLP:conf/icmcs/WuLLWYL19}, \cite{DBLP:journals/corr/abs-1807-11334}, \cite{fu2019self}. 
Nevertheless, the optimal performance is often achieved by ensemble that integrates multiple sub-networks and their discrimination capability. However, ensemble learning in domain adaptive re-ID remains unexplored. How to leverage specific features and knowledge of multiple networks and optimally adapt them to an unlabelled target domain remains to be elaborated.

In this paper, we present an multiple expert brainstorming network (MEB-Net), which learns and adapts multiple networks with different architectures for optimal re-ID in an unlabelled target domain. 
MEB-Net conducts iterative training where clustering for pseudo-labels and models feature learning are alternately executed. 
For feature learning, MEB-Net adopts a mutual learning strategy where networks with different architectures are pre-trained in a source domain as expert models equipped with specific features and  knowledge.
The adaptation is accomplished through brainstorming-based mutual learning among multiple expert models. 
To accommodate the heterogeneity of experts learned with different architectures, a regularization scheme is introduced to modulate the experts' authority according to their feature distributions in the target domain, and further enhances the discrimination capability of the re-ID model.

The contributions of this paper are summarized as follows.
\begin{itemize}
    \item  We propose a novel multiple expert brainstorming network (MEB-Net) based on mutual learning among expert models, each of which is equipped with knowledge of an architecture.  
    
    \item We design an authority regularization to accommodate the heterogeneity of experts learned with different architectures, modulating the authority of experts and enhance the discrimination capability of re-ID models.
    
    \item Our MEB-Net approach achieves significant performance gain over the state-of-the-art on commonly used datasets: Market-1501 and DukeMTMC-reID.
\end{itemize}  

\section{Related Works}

\subsection{Unsupervised Domain Adaptive Re-ID}

Unsupervised domain adaptation (UDA) for person re-ID defines a learning problem for target domains where source domains are fully labeled while sample labels in target domains are totally unknown. Methods have been extensively explored in recent years, which take three typical approaches as follows.

\textbf{Feature distribution alignment.} 
In~\cite{DBLP:conf/bmvc/LinLLK18}, Lin \emph{et al.} proposed minimizing the distribution variation of the source's and the target's mid-level features based on Maximum Mean Discrepancy (MMD) distance. Wang \emph{et al.} \cite{Wang_2018_CVPR} utilized additional attribute annotations to align feature distributions of source and target domains in a common space. 

\textbf{Image-style transformation.}
GAN-based methods have been extensively explored for domain adaptive person re-ID \cite{DBLP:conf/ksem/LvW18}, \cite{DBLP:conf/eccv/ZhongZLY18}, \cite{Wei_2018_CVPR}, \cite{Deng_2018_CVPR}, \cite{Liu_2019_CVPR}. 
HHL \cite{DBLP:conf/eccv/ZhongZLY18} simultaneously enforced cameras invariance and domain connectedness to improve the generalization
ability of models on the target set. PTGAN \cite{Wei_2018_CVPR}, SPGAN \cite{Deng_2018_CVPR}, ATNet \cite{Liu_2019_CVPR} and PDA-Net \cite{li2019cross} transferred images with identity labels from source into target domains to learn discriminative models.

\textbf{Self-supervised learning.}
Recently, the problem about how to leverage the large number of unlabeled samples in target domains have attracted increasing attention \cite{DBLP:journals/corr/FanZY17}, \cite{DBLP:conf/iccv/YeMZLY17}, \cite{DBLP:conf/iccv/LiuWL17}, \cite{DBLP:conf/icmcs/WuLLWYL19}, \cite{Wu_2018_CVPR}, \cite{Zhong_2019_CVPR}, \cite{zhang2019self}. Clustering~\cite{DBLP:journals/corr/FanZY17}, \cite{DBLP:conf/eccv/ZhengBSWSWT16}, \cite{Zhai_2020_CVPR} and graph matching~\cite{DBLP:conf/iccv/YeMZLY17} methods have been explored to predict pseudo-labels in target domains for discriminative model learning. Reciprocal search~\cite{DBLP:conf/iccv/LiuWL17} and exemplar-invariance approaches~\cite{Wu_2018_CVPR} were proposed to refine pseudo labels, taking camera-invariance into account concurrently.  SSG \cite{fu2019self} utilized both global and local feature of persons to build multiple clusters, which are then assigned pseudo-labels to supervise the model training. 

However, existing works barely explored the domain adaptive person re-ID task using methods of model ensemble, which have achieved impressive performance on many other tasks.  

\subsection{Knowledge Transfer}
Distilling knowledge from well trained neural networks and transferring it to another model/network has been widely studied in recent years \cite{hinton2015distilling}, \cite{chen2015net2net}, \cite{li2017learning}, \cite{yim2017gift}, \cite{bagherinezhad2018label}, \cite{anil2018large}. The typical approach of knowledge transfer is the teacher-student model learning, which uses the soft output distribution of a teacher network to supervise a student network, so as to make student models learn discrimination ability from teacher models.

The mean-teacher model~\cite{tarvainen2017mean} averaged model weights at different training iterations to create supervisions for unlabeled samples. Deep mutual learning~\cite{zhang2018deep} adopted a pool of student models by training them with supervision from each other. Mutual mean teaching~\cite{ge2020mutual} designed a symmetrical framework with hard pseudo-labels as well as refined soft labels for unsupervised domain adaptive re-ID. However, existing methods with teacher-student mechanisms mostly adopted a symmetrical framework which largely neglected the different confidence of teacher networks when they are heterogeneous.

\subsection{Model Ensemble}
There is a considerable number of previous works on ensembles with neural networks. A typical approach \cite{srivastava2014dropout}, \cite{wan2013regularization}, \cite{huang2016deep}, \cite{singh2016swapout} generally create a series of
networks with shared weights during training and then implicitly
ensemble them at test time. Another approach \cite{shen2019meal} focus on label refinery by well trained networks for training a new model with higher discrimination capability. However, these methods cannot be directly used on unsupervised domain adaptive re-ID tasks, where the training set and the testing set share non-overlapping label space.

\begin{figure}[t]
\centering
\includegraphics[width=1.0\linewidth]{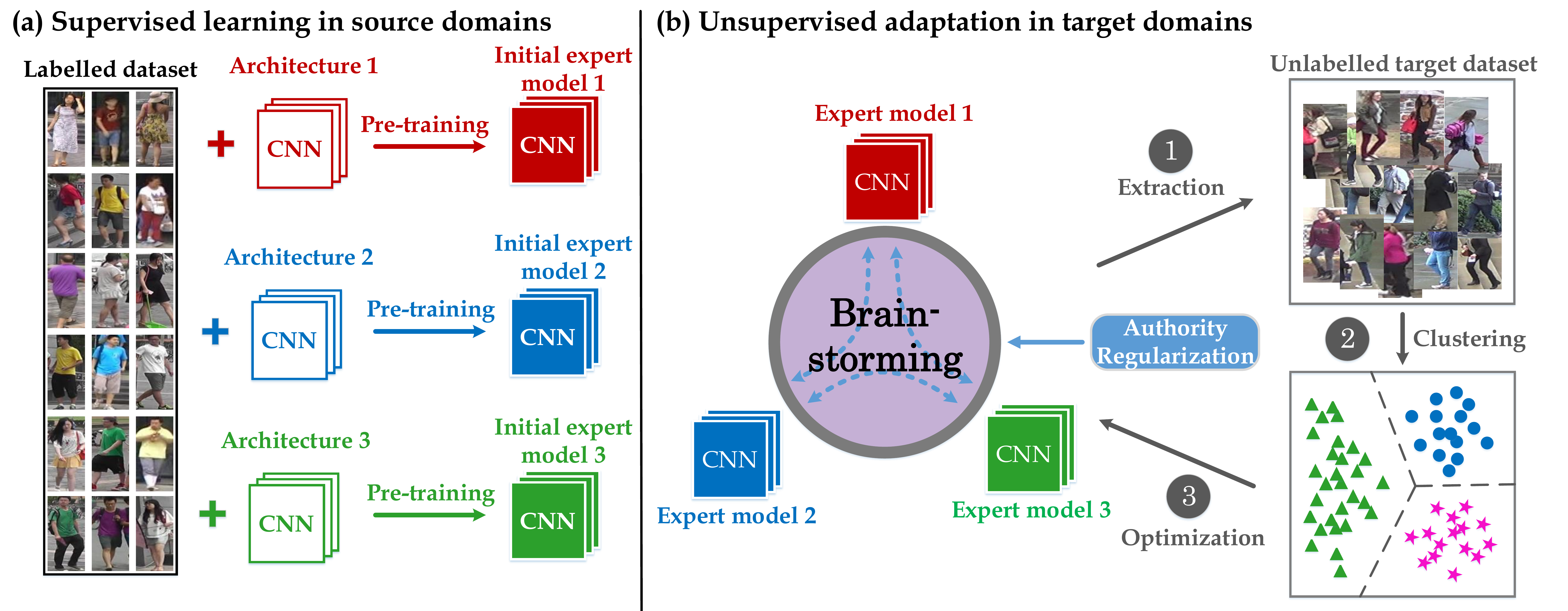}

\caption{Overview of proposed multiple expert brainstorming network (MEB-Net). Multiple expert networks with different architectures are first pre-trained in the source domain and then adapted to the target domain through brainstorming.}
\label{fig:overview}

\end{figure}

\section{The Proposed Approach}

\begin{figure}[t]
\centering
\includegraphics[width=1.0\linewidth]{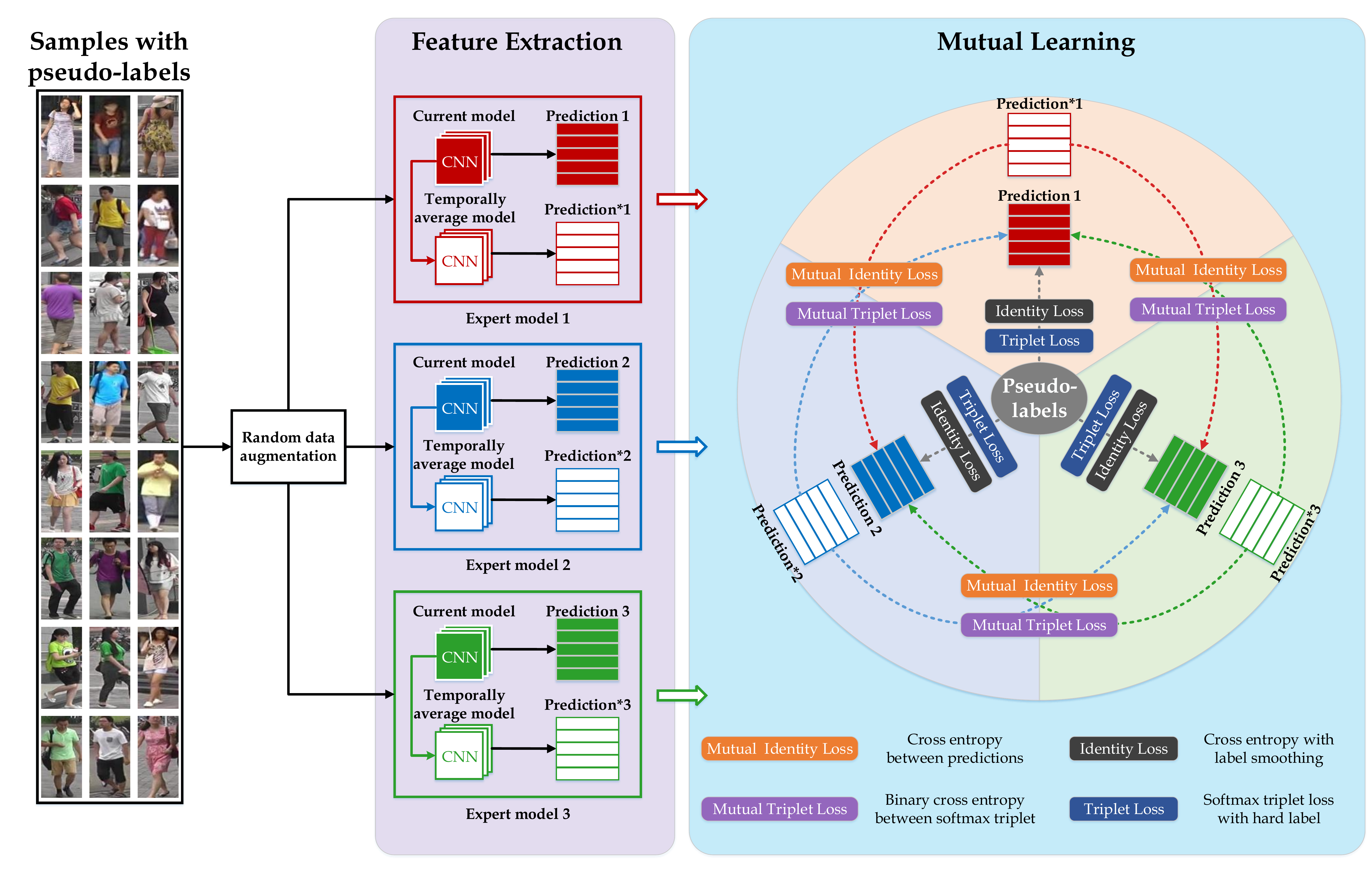}

\caption{Flowchart of proposed expert brainstorming in MEB-Net, which consists of two components, feature extraction and mutual learning. 
In mutual learning, multiple expert networks are organized to collaboratively learn from each other by their predictions and the pseudo-labels, and improve themselves for the target domain in an unsupervised mutual learning manner. 
More details are described in Sec. \ref{sec:bs}.}
\label{fig:flowchart}

\end{figure}

We study the problem of unsupervised domain adaptive re-ID using model ensemble methods from a source-domain to a target-domain.
%
The labelled source-domain dataset are denoted as $\mathcal{S}=\{X_s, Y_s\}$, which has $N_s$ sample images with $M_s$ unique identities.
$X_s$ and $Y_s$ denote the sample images and the person identities, where each sample $x_s$ in $X_s$ is associated with a person identity $y_s$ in $Y_s$.
%
%
The $N_t$ sample images in the target-domain $\mathcal{T} = \{X_t\}$ have no identity available. 
We aim to leverage the labelled sample images in $\mathcal{S}$ and the unlabelled sample images in $\mathcal{T}$ to learn a transferred re-ID model for the target-domain $\mathcal{T}$.

\subsection{Overview}


MEB-Net adopts a two-stage training scheme including supervised learning in source domains (Fig.\ \ref{fig:overview}a) and unsupervised adaptation to target domains (Fig.\ \ref{fig:overview}b). 
In the initialization phase, multiple expert models with different network architectures are pre-trained by the source dataset in a supervised manner. 
Afterwards the trained experts are adapted to the target domain by iteratively brainstorming with each other using the unlabelled target-domain samples.
In each iterative epoch, pseudo-labels are predicted for target samples via clustering which are then utilized to fine-tune the expert networks by mutual learning.
In addition, the authority regularization is employed to modulate the authority of expert networks according to their discrimination capability during training. 
In this way, the knowledge from multiple networks is fused, enhanced, and transferred to the target domain, as described in \textbf{Algorithm \ref{alg:algorithm}}.
%

\subsection{Learning in Source Domains}
\label{sec:pretrain}
The proposed MEB-Net aims to transfer the knowledge of multiple networks from a labelled source domain to an unlabelled target domain. 
For each architecture, a deep neural network (DNN) model $\mathcal{M}^k$  parameterized with $\theta^k$ (a pre-trained expert) is first trained in a supervised manner. 
$\mathcal{M}^k$ transforms each sample image $x_{s,i}$ into a feature representation $f(x_{s,i}|\theta^k)$, and outputs a predicted probability $p_j(x_{s,i}|\theta^k)$ of image $x_{s,i}$ belonging to the identity $j$. 
The cross entropy loss with label smoothing is defined as
\begin{equation}
\begin{aligned}
    \mathcal{L}_{s,id}^k = \frac{1}{N_s} \sum_{i=1}^{N_s} \sum_{j=1}^{M_s} q_j \log p_j(x_{s,i}|\theta^k) 
\end{aligned}
\end{equation}
where $q_j = 1-\varepsilon + \frac{\varepsilon}{M_s}$ if $j=y_{s,i}$, otherwise $q_j=\frac{\varepsilon}{M_s}$. $\varepsilon$ is a small constant, which is set as 0.1. The softmax triplet loss is also defined as
\begin{equation}
\begin{aligned}
    \mathcal{L}_{s,tri}^k =- \frac{1}{N_s} \sum_{i=1}^{N_s}  \log \frac{e^{\|f(x_{s,i}|\theta^k)-f(x_{s,i-}|\theta^k)\|}}{e^{\|f(x_{s,i}|\theta^k)-f(x_{s,i+}|\theta^k)\|}+e^{\|f(x_{s,i}|\theta^k)-f(x_{s,i-}|\theta^k)\|} }
\end{aligned}
\end{equation}
where $x_{s,i+}$ denotes the hardest positive sample of the anchor $x_{s,i}$, and $x_{s,i-}$ denotes the hardest negative sample. $\|\cdot\|$ denotes the L$_2$ distance. The overall loss is therefore calculated as
\begin{equation}
\begin{aligned}
    \mathcal{L}_{s}^k = \mathcal{L}_{s,id}^k + \mathcal{L}_{s,tri}^k.
\end{aligned}
\end{equation}

With $K$ network architectures, the supervised learning thus produces $K$ pre-trained re-ID models each of which acts as an expert for brainstorming.

\begin{algorithm}[t]
\small
\caption{Multiple Expert Brainstorming Network}
\label{alg:algorithm}
\textbf{Input}:  Source domain dataset $\mathcal{S}=\{X_s, Y_s\}$, target domain dataset $\mathcal{T}=\{X_t\}$.\\
\textbf{Input}:  K network architectures $\{\mathcal{A}^k\}$.\\
\textbf{Output}: Expert model parameters $\{\theta^k\}$.
\begin{algorithmic}[1] 
\STATE Initialize pre-trained weights $\theta^k$ of model $\mathcal{M}^k$ with each architecture $\mathcal{A}^k$. 
\FOR{each epoch}
\STATE Extract average features on $\mathcal{T}$: $f(X_t) = \frac{1}{K} \sum_{k=1}^K f(X_t|\Theta^k )$.
\STATE Generate pseudo-labels $\widetilde{Y}_t$ of $X_t$ by clustering samples using $f(X_t)$.
\STATE Evaluate authority $w$ of each expert model by inter-/intra-cluster scatter.
\FOR{each iteration $T$, mini-batch $\mathcal{B}\subset \mathcal{T}$ }
\STATE Calculate soft-labels from each temporally average model with $\{\Theta_{T}^k\}$: $p(x_{t,i\in \mathcal{B}}|\Theta_{T}^k)$, $\mathcal{P}_{i\in \mathcal{B}}({\Theta}_T^k)$.
\STATE Calculate output of each current model with $\{\theta^k\}$: $p(x_{t,i\in \mathcal{B}}|\theta^k)$, $\mathcal{P}_{i\in \mathcal{B}}({\theta}^k)$.
\STATE Update parameters $\{\theta^k\}$ by optimizing Eq. \ref{eq:overall} with authority $\{w^e\}$.
\STATE Update temporally average model weights $\{\Theta_{T}^k\}$ following Eq.~\ref{eq:average}.
\ENDFOR
\ENDFOR
\STATE \textbf{Return} Expert model parameters $\{\theta^k\}$
\end{algorithmic}
\end{algorithm}

\subsection{Clustering in the Target Domain}

\label{sec:cluster}
In the target domain, MEB-Net consists of a clustering-based pseudo-label generation procedure and a feature learning procedure, which are mutually enforced. 
Each epoch consists of three steps: (1) For sample images in the target domain, each expert model extracts convolutional features $f(X_t|\theta^k)$ and determines the ensemble features by averaging features extracted by multiple expert models $f(X_t) = \frac{1}{K} \sum_{k=1}^K f(X_t|\theta^k)$; 
(2) A mini-batch k-means clustering is performed on $f(X_t)$ to classify all target-domain samples into $M_t$ different clusters; 
(3) The produced cluster IDs are used as pseudo-labels $\widetilde{Y_t}$ for the training samples $X_t$. The steps 3 and 4 in \textbf{Algorithm \ref{alg:algorithm}} summarize this clustering process.

\subsection{Expert Brainstorming}
\label{sec:bs}
With multiple expert models $\{\mathcal{M}^k\}$ with different architectures which absorb rich knowledge from the source domain, MEB-Net aims to organize them to collaboratively learn from each other and improve themselves for the target domain in an unsupervised mutual learning manner, Fig. \ref{fig:flowchart}. 

In each training iteration, the same batch of images in the target domain are first fed to all the expert models $\{\mathcal{M}^k\}$ parameterized by $\{\theta^k\}$, to predict the classification confidence predictions $\{p(x_{t,i}|\theta^k)\}$ and feature representations $\{f(x_{t,i}|\theta^k)\}$.
To transfer knowledge from one expert to others, the class predictions of each expert can serve as soft class labels for training other experts. 
However, directly using the current predictions as soft labels to train each model decreases the independence of expert models' outputs, which might result in an error amplification.
To avoid this error, MEB-Net leverages the temporally average model of each expert model, which preserves more original knowledge, to generate reliable soft pseudo labels for supervising other experts. The parameters of the temporally average model of expert $\mathcal{M}^k$ at current iteration $T$ are denoted as $\Theta_T^k$, which is updated as
\begin{equation}
\label{eq:average}
\begin{aligned}
    {\Theta}_{T}^k = \alpha {\Theta}_{T-1}^k + (1-\alpha)\theta^k,
\end{aligned}
\end{equation}
where $\alpha \in [0,1]$ is the scale factor, and the initial temporal average parameters are ${\Theta}_{0}^k = \theta^k$. Utilizing this temporal average model of expert $\mathcal{M}^e$, the probability for each identity $j$ is predicted as $p_j(x_{t,i}|\Theta_{T}^e)$, and the feature representation is calculated as $f(x_{t,i}|\Theta_{T}^e)$.

\textbf{Mutual identity loss.}
For each expert model $\mathcal{M}^k$, the mutual identity loss of models learned by a certain expert $\mathcal{M}^e$ is defined as the cross entropy between the class prediction of the expert $\mathcal{M}^k$ and the temporal average model of the expert $\mathcal{M}^e$, as
\begin{equation}
\begin{aligned}
    \mathcal{L}_{mid}^{k\leftarrow e} = - \frac{1}{N_t} \sum_{i=1}^{N_t}  \sum_{j=1}^{M_t} p_j(x_{t,i}|\Theta_{T}^e) \log p_j(x_{t,i}|\theta^k).
\end{aligned}
\end{equation}
The mutual identity loss for expert $\mathcal{M}^k$ is set as the average of above losses of models learned by all other experts, as
\begin{equation}
\begin{aligned}
    \mathcal{L}_{mid}^k = \frac{1}{K-1} \sum_{e\neq k}^{K} \mathcal{L}_{mid}^{k\leftarrow e}
\end{aligned}
\label{eq:mid}
\end{equation}

\textbf{Mutual triplet loss.}
For each expert model $\mathcal{M}^k$, the mutual triplet loss of models learned by a certain expert $\mathcal{M}^e$ is also defined as binary cross entropy, as
\begin{equation}
\begin{aligned}
    \mathcal{L}_{mtri}^{k\leftarrow e} = -\frac{1}{N_t} \sum_{i=1}^{N_t} \bigg[ \mathcal{P}_i({\Theta}_{T}^e) \log \mathcal{P}_i({\theta}^k) + (1-\mathcal{P}_i({\Theta}_{T}^e)) \log (1-\mathcal{P}_i({\theta}^k)) \bigg],
\end{aligned}
\end{equation}
where $\mathcal{P}_i({\theta}^k)$ denotes the softmax of the feature distance between negative sample pairs:
\begin{equation}
\begin{aligned}
    \mathcal{P}_i({\theta}^k) = \frac{e^{\|f(x_{t,i}|\theta^k)-f(x_{t,i-}|\theta^k)\|}}{e^{\|f(x_{t,i}|\theta^k)-f(x_{t,i+}|\theta^k)\|}+e^{\|f(x_{t,i}|\theta^k)-f(x_{t,i-}|\theta^k)\|} },
\end{aligned}
\end{equation}
where $x_{t,i+}$ denotes the hardest positive sample of the anchor $x_{t,i}$ according to the pseudo-labels $\widetilde{Y_t}$, and $x_{t,i-}$ denotes the hardest negative sample. $\|\cdot\|$ denotes $L_2$ distance.
The mutual triplet loss for expert $\mathcal{M}^k$ is calculated as the average of above triplet losses of models learned by all other experts, as
\begin{equation}
\begin{aligned}
    \mathcal{L}_{mtri}^k = \frac{1}{K-1} \sum_{e\neq k}^{K} \mathcal{L}_{mtri}^{k\leftarrow e},
\end{aligned}
\label{eq:mtri}
\end{equation}

\textbf{Voting loss.} In order to learn stable and discriminative knowledge from the pseudo-labels obtained by clustering as described in Sec. \ref{sec:cluster}, we introduce voting loss which consists of the identity loss and the triplet loss. For each expert model $\mathcal{M}^k$, the identity loss is defined as cross entropy with label smoothing, as
\begin{equation}
\begin{aligned}
    \mathcal{L}_{id}^k = \frac{1}{N_t} \sum_{i=1}^{N_t} \sum_{j=1}^{M_t} q_j \log p_j(x_{t,i}|\theta^k),
\end{aligned}
\end{equation}
where $q_j = 1-\varepsilon + \frac{\varepsilon}{M_t}$ if $j=\widetilde{y}_{t,i}$, otherwise $q_j=\frac{\varepsilon}{M_t}$. $\varepsilon$ is small constant. The softmax triplet loss is defined as:
\begin{equation}
\begin{aligned}
    \mathcal{L}_{tri}^k =- \frac{1}{N_t} \sum_{i=1}^{N_t}  \log \frac{e^{\|f(x_{t,i}|\theta^k)-f(x_{t,i-}|\theta^k)\|}}{e^{\|f(x_{t,i}|\theta^k)-f(x_{t,i+}|\theta^k)\|}+e^{\|f(x_{t,i}|\theta^k)-f(x_{t,i-}|\theta^k)\|} },
\end{aligned}
\end{equation}
where $x_{t,i+}$ denotes the hardest positive sample of the anchor $x_{t,i}$, and $x_{t,i-}$ denotes the hardest negative sample. $\|\cdot\|$ denotes L$_2$ distance. The voting loss is defined by summarizing the identity loss and the triplet loss:
\begin{equation}
\label{eq:vot}
\begin{aligned}
    \mathcal{L}_{vot}^k = \mathcal{L}_{id}^k + \mathcal{L}_{tri}^k,
\end{aligned}
\end{equation}

\textbf{Overall loss.} For each expert model $\mathcal{M}^k$, the individual brainstorming loss is defined by
\begin{equation}
\begin{aligned}
    \mathcal{L}_{bs}^k = \mathcal{L}_{mid}^k + \mathcal{L}_{mtri}^k + \mathcal{L}_{vot}^k,
\end{aligned}
\end{equation}
The overall loss is defined by the sum loss of the individual brainstorming for each expert model.
\begin{equation}
\label{eq:overall}
\begin{aligned}
    \mathcal{L}_{meb} = \sum_{k=1}^K \mathcal{L}_{bs}^{k}.
\end{aligned}
\end{equation}

\subsection{Authority Regularization}
\label{sec:ar}

\begin{figure}[t]
\centering
\includegraphics[width=1.0\linewidth]{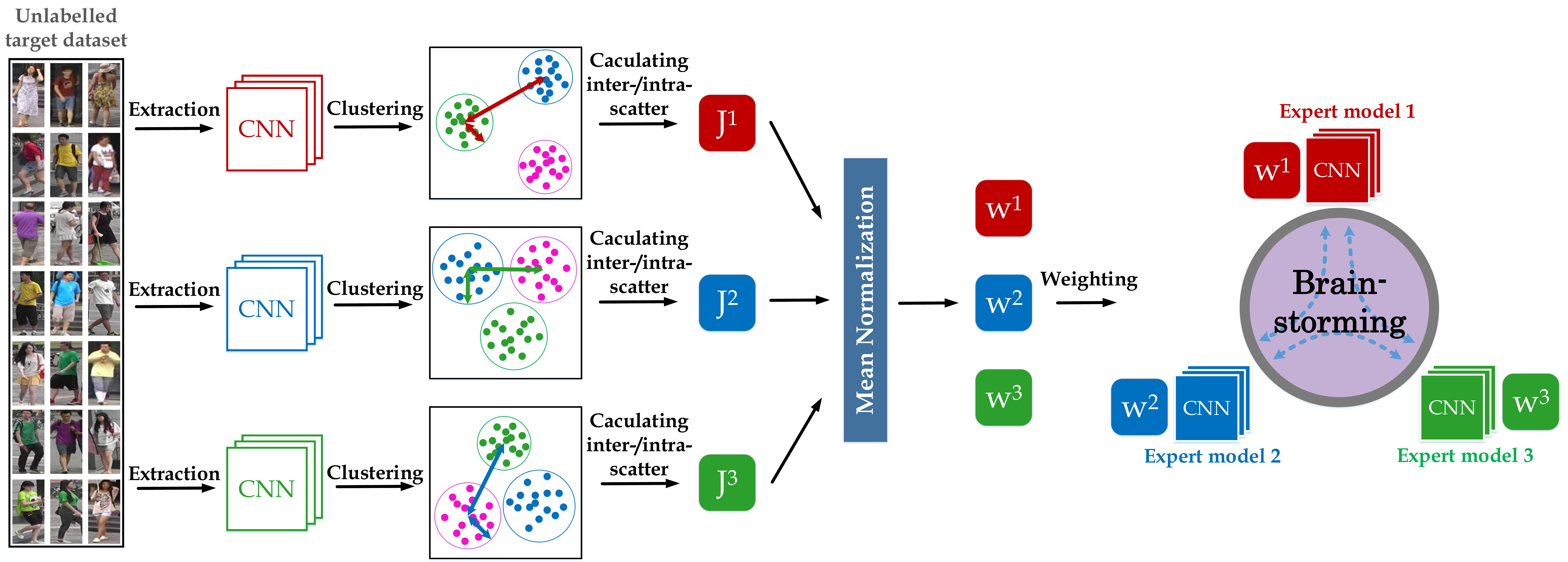}
\caption{Illustration of our proposed authority regularization. It modulates the authority of different experts according to the inter-/intra-cluster scatter of each single expert. A larger scatter means better discrimination capability.}
\label{fig:ar}
\end{figure}

Expert networks with different architectures are equipped with various knowledge, and thus have different degrees of discrimination capability in the target domain. 
To accommodate the heterogeneity of experts, we propose an authority regularization (AR) scheme, which
modulates the authority of different experts according to the inter-/intra-cluster scatter of each single expert, Fig. \ref{fig:ar}. 
Specifically, for each expert $\mathcal{M}$ we extract sample features $f(x|\Theta_T)$ and cluster all the training samples in the target domain into $M_t$ groups as $\mathbb{C}$. 
The intra-cluster scatter of the cluster $\mathbb{C}_i$ is defined as
\begin{equation}
\begin{aligned}
    S_{intra}^i = \sum_{x \in  \mathbb{C}_i}  \|f(x|\Theta_T) - \mu_i \|^2,
\end{aligned}
\end{equation}
where $\mu_i=\sum_{x \in  \mathbb{C}_i} f(x|\Theta_T)/n_t^i$ is the average feature of the cluster $\mathbb{C}_i$ (with $n_t^i$ samples). The inter-cluster scatter is defined as
\begin{equation}
\begin{aligned}
    S_{inter} = \sum_{i=1}^{M_t} n_i^t \|\mu_i - \mu\|^2,
\end{aligned}
\end{equation}
where $\mu=\sum_{i=1}^{N_t} f(x_{t,i}|\Theta_T)/N_t$ is the average feature of all training samples in the target domain. To evaluate the discrimination of each expert model in the unlabeled target domain, the inter-/intra-cluster scatter $J$ is defined as
\begin{equation}
\begin{aligned}
    J = \frac{S_{inter}}{\sum_{i=1}^{M_t} S_{intra}^i}.
\end{aligned}
\end{equation}
$J$ gets larger when the inter-cluster scatter is larger or the intra-cluster scatter is smaller. And a larger $J$ means better discrimination capability. Before feature learning in each epoch, we calculate $J$ scatter for each expert $\mathcal{M}^e$ as $J^e$, and defined expert authority $w^e$ as the mean normalization of $J^e$, as
\begin{equation}
\begin{aligned}
    w^e = \frac{J^e}{\sum_{k=1}^K J^k/K} = \frac{K J^e}{\sum_{k=1}^K J^k}.
\end{aligned}
\end{equation}
We re-define the mutual identity loss in Eq. \ref{eq:mid} and the mutual triplet loss in Eq. \ref{eq:mtri} as the weighted sum of $\mathcal{L}_{mid}^{k\leftarrow e}$ and $\mathcal{L}_{mtri}^{k\leftarrow e}$ for other experts, as
\begin{equation}
\label{eq:arid}
\begin{aligned}
    \mathcal{L}_{mid}^k = \frac{1}{K-1} \sum_{e\neq k}^{K} w^e \mathcal{L}_{mid}^{k\leftarrow e},
\end{aligned}
\end{equation}
and
\begin{equation}
\label{eq:artri}
\begin{aligned}
    \mathcal{L}_{mtri}^k = \frac{1}{K-1} \sum_{e\neq k}^{K} w^e \mathcal{L}_{mtri}^{k\leftarrow e}.
\end{aligned}
\end{equation}
With the regularization scheme, MEB-Net modulates the authority of experts to facilitate discrimination in the target domain.




\section{Experiments}

\subsection{Datasets and Evaluation Metrics}
We evaluate the proposed method on  Market-1501~\cite{Zheng_2015_ICCV} and DukeMTMC-reID~\cite{DBLP:conf/eccv/RistaniSZCT16}\cite{Zheng_2017_ICCV}.

\textbf{Market-1501:} 
This dataset contains 32,668 images of 1,501 identities from 6 disjoint cameras, among which 12,936 images from 751 identities form a training set, 19,732 images from 750 identities (plus a number of distractors) form a gallery set, and 3,368 images from 750 identities form a query set.

\textbf{DukeMTMC-reID:} This dataset is a subset of the DukeMTMC. It consists of 16,522 training images,
2,228 query images, and 17,661 gallery images of 1,812 identities captured using 8 cameras. Of the 1812 identities,
1,404 appear in at least two cameras and the rest (distractors) appear in a single camera.



\textbf{Evaluation Metrics:} 
In evaluations, we use one dataset as the target domain and the other as the source domain. 
The used metrics are Cumulative Matching Characteristic (CMC) curve and mean average precision (mAP).

\subsection{Implementation Details}
MEB-Net is trained by two stages: \textit{pre-training in source domains} and the \textit{adaptation in target domains}.

\textbf{Stage 1: Pre-training in source domains:}
We first pre-train three supervised expert models on the source dataset as described in Section~\ref{sec:pretrain}. 
We adopt three architectures: DenseNet-121 \cite{huang2017densely}, ResNet-50 \cite{He_2016_CVPR} and Inception-v3 \cite{szegedy2016rethinking} as backbone networks for the three experts, and initialize them by using parameters pre-trained on the ImageNet \cite{DBLP:conf/cvpr/DengDSLL009}. Zero padding is employed on the final features to obtain representations of the same 2048 dimensions for all networks.
During training, the input image is resized to $256\times128$ and traditional image augmentation was performed via random flipping and random erasing.
%
For each identity from the training set, a mini-batch of 64 is sampled with $P$ = 16 randomly selected identities and $K$ = 4 randomly sampled images for computing the hard batch triplet loss. We use the Adam \cite{kingma2014adam} with weight decay 0.0005 to optimize parameters. The initial learning rate is set to 0.00035 and is decreased to 1/10 of its previous value on the 40th and 70th epoch in the total 80 epochs.

\textbf{Stage 2: Adaptation in target domains.}
For unsupervised adaptation on target datasets, we follow the same data augmentation strategy and triplet loss setting. The temporal ensemble momentum $\alpha$ in Eq \ref{eq:average} is set to 0.999. The learning rate is fixed to 0.00035 for overall 40 epochs. In each epoch, we conduct mini-batch k-means clustering and the number of groups $M_t$ is set as 500 for all target datasets. Each epoch consists of 800 training iterations. During testing, we only use one expert network for feature representations.

\subsection{Comparison with State-of-the-Arts}

\begin{table*}[t]
\begin{center}
\begin{tabular}{l|p{1.0cm}<{\centering}p{1.0cm}<{\centering}p{1.0cm}<{\centering}p{1.0cm}<{\centering}|p{1.0cm}<{\centering}p{1.0cm}<{\centering}p{1.0cm}<{\centering}p{1.0cm}<{\centering}}
    \hline\hline
    \multirow{2}{*}{Methods} & \multicolumn{4}{c|}{Market-1501} & \multicolumn{4}{c}{DukeMTMC-reID} \\
    \cline{2-9}
    & mAP & R-1 & R-5 & R-10 & mAP & R-1 & R-5 & R-10  \\
    \hline\hline
    LOMO\cite{Liao_2015_CVPR}             & 8.0 & 27.2 & 41.6 & 49.1   & 4.8  & 12.3 & 21.3 & 26.6  \\
    Bow\cite{Zheng_2015_ICCV}               & 14.8 & 35.8 & 52.4 & 60.3  & 8.3  & 17.1 & 28.8 & 34.9  \\
    UMDL\cite{Peng_2016_CVPR}              & 12.4 & 34.5 & 52.6 & 59.6  & 7.3 & 18.5 & 31.4 & 37.6  \\
    \hline
    MMFA\cite{DBLP:conf/bmvc/LinLLK18}   & 27.4   & 56.7 & 75.0 & 81.8 & 24.7 & 45.3 & 59.8 & 66.3 \\
    TJ-AIDL\cite{Wang_2018_CVPR}            & 26.5 & 58.2 & 74.8 & 81.1 & 23.0 & 44.3 & 59.6 & 65.0 \\
    UCDA-CCE\cite{qi2019novel} & 30.9 & 60.4 & - & - & 31.0 & 47.7 & - & - \\
    \hline
    ATNet\cite{Liu_2019_CVPR}              & 25.6 & 55.7 & 73.2 & 79.4  & 24.9 & 45.1 & 59.5 & 64.2 \\
    SPGAN+LMP\cite{Deng_2018_CVPR}         & 26.7  & 57.7 & 75.8 & 82.4 & 26.2 & 46.4 & 62.3 & 68.0  \\
    CamStyle\cite{DBLP:journals/tip/ZhongZZLY19} & 27.4  & 58.8 & 78.2 & 84.3 & 25.1 & 48.4 & 62.5 & 68.9 \\
    HHL\cite{DBLP:conf/eccv/ZhongZLY18}   & 31.4  & 62.2 & 78.8 & 84.0  & 27.2 & 46.9 & 61.0 & 66.7 \\
    ECN\cite{Zhong_2019_CVPR}      & 43.0 & 75.1 & 87.6 & 91.6 &  40.4 & 63.3 & 75.8 & 80.4  \\
    PDA-Net\cite{li2019cross} & 47.6 & 75.2 & 86.3 & 90.2 & 45.1 & 63.2 & 77.0 & 82.5 \\
    \hline
    PUL\cite{DBLP:journals/tomccap/FanZYY18}   & 20.5   & 45.5 & 60.7 & 66.7 & 16.4 & 30.0 & 43.4 & 48.5\\
    UDAP\cite{DBLP:journals/corr/abs-1807-11334}   & 53.7  & 75.8 & 89.5 & 93.2  & 49.0 & 68.4 & 80.1 & 83.5 \\
    PCB-PAST\cite{zhang2019self} & 54.6 & 78.4 & - & - & 54.3 & 72.4 & - & - \\
    SSG\cite{fu2019self} & \underline{58.3} & \underline{80.0} & \underline{90.0} & 
                        \underline{92.4} &  \underline{53.4} &
                        \underline{73.0} & \underline{80.6} & 
                        \underline{83.2}  \\
    MMT-500\cite{ge2020mutual}     & \emph{71.2} & \emph{87.7} & 
                        \emph{94.9}   & \emph{96.9} & \emph{63.1} & 
                        \emph{76.8} & \emph{88.0} & 
                        \emph{92.2} \\
    \hline
    MEB-Net(Ours) & \textbf{76.0} & \textbf{89.9} & 
                        \textbf{96.0} & \textbf{97.5} & 
                        \textbf{66.1} & \textbf{79.6} & 
                        \textbf{88.3} & \textbf{92.2} \\
\hline\hline
\end{tabular}
\end{center}
\caption{Comparison with state-of-the-art methods: For the adaptation on Market-1501 and that on DukeMTMC-reID. The top-three results are highlighted with bold, italic, and underline fonts, respectively.}
\label{table:staart}
\end{table*}

We compare MEB-Net with state-of-the-art methods including: hand-crafted feature approaches (LOMO\cite{Liao_2015_CVPR}, BOW\cite{Zheng_2015_ICCV}, UMDL\cite{Peng_2016_CVPR}), feature alignment based methods (MMFA\cite{DBLP:conf/bmvc/LinLLK18}, TJ-AIDL\cite{Wang_2018_CVPR}, UCDA-CCE\cite{qi2019novel}), GAN-based methods (SPGAN~\cite{Deng_2018_CVPR}, ATNet\cite{Liu_2019_CVPR}, CamStyle\cite{DBLP:journals/tip/ZhongZZLY19}, HHL\cite{DBLP:conf/eccv/ZhongZLY18}, ECN\cite{Zhong_2019_CVPR} and PDA-Net\cite{li2019cross}), pseudo-label prediction based methods (PUL\cite{DBLP:journals/tomccap/FanZYY18}, UDAP\cite{DBLP:journals/corr/abs-1807-11334}, PCB-PAST\cite{zhang2019self}, SSG\cite{fu2019self} MMT\cite{ge2020mutual}). Table \ref{table:staart} shows the person Re-ID performance while adapting from Market1501 to DukeMTMC-reID and vice versa.

\textbf{Hand-crafted feature approaches.} As Table.\ref{table:staart} shows, MEB-Net outperforms hand-crafted feature approaches including LOMO, BOW and UMDL by large margins, as deep network can learn more discriminative representations than hand-crafted features. 

\textbf{Feature alignment approaches.} MEB-Net significantly exceeds the feature alignment unsupervised Re-ID models. The reason lies in that it explores and utilizes the similarity between unlabelled sample in target domains in an more effective manner of brainstorming.

\textbf{GAN-based approaches.} The performance of these approaches is diverse. In particular, ECN performs better than most methods using GANs because it enforces cameras in-variance as well as latent sample relations. However, MEB-Net can achieve higher performance than GAN-based methods without generating new images, which indicates its more efficient use of the unlabelled samples. 

\textbf{Pseudo-labels based approaches.} The line of approaches perform clearly better than other approaches in most cases, as they fully make use of the unlabelled target samples by assigning pseudo-labels to them according to sample feature similarities. 
For a fair comparison, we report MMT-500 with the cluster number of 500, which is the same as the proposed MEB-Net. As Table.\ref{table:staart} shows, MEB-Net achieves an mAP of $76.0\%$ and a rank-1 accuracy of $89.9\%$ for the DukeMTMC-reID$\rightarrow$Market1501 transfer, which outperforms the state-of-the-art (by MMT-500) by $4.8\%$ and $2.2\%$, respectively. And for Market1501$\rightarrow$DukeMTMC-reID transfer, MEB-Net obtains an mAP of $66.1\%$ and a rank-1 accuracy of $79.6\%$ which outperforms the state-of-the-art by $3.0\%$ and $2.8\%$, respectively.

\subsection{Ablation Studies}

\begin{table*}[t]
\begin{center}
\begin{tabular}{l|p{0.9cm}<{\centering}p{0.9cm}<{\centering}p{0.9cm}<{\centering}p{0.9cm}<{\centering}|p{0.9cm}<{\centering}p{0.9cm}<{\centering}p{1.0cm}<{\centering}p{1.0cm}<{\centering}}
    \hline\hline
    \multirow{2}{*}{Methods} & \multicolumn{4}{c|}{Market-1501} & \multicolumn{4}{c}{DukeMTMC-reID} \\
    \cline{2-9}
    & mAP & R-1 & R-5 & R-10 & mAP & R-1 & R-5 & R-10 \\
    \hline\hline
    Supervised Models  & 82.5 & 93.7 & 98.1 & 98.5  & 67.1 & 82.1 & 90.0 & 92.1  \\
    \hline
    Direct Transfer & 31.5 & 60.6 & 75.7 & 80.8  & 29.7 & 46.5 & 61.8 & 67.7  \\
    Baseline(Only $\mathcal{L}_{vot}$)  & 69.5 & 86.8 & 94.9 & 96.6 & 60.6 & 75.0 & 85.5 & 89.4  \\
    MEB-Net w/o $\Theta_T$            &  70.7 & 87.1  & 94.8  & 96.7  & 58.3 & 72.6  & 83.6  & 88.5  \\
    MEB-Net w/o $\mathcal{L}_{mid}$    &  70.2 & 87.9  & 94.8  & 96.6  & 60.4 & 75.0  & 86.1  & 89.3   \\
    MEB-Net w/o $\mathcal{L}_{mtri}$    & 74.9  & 88.4  & 95.8  & 97.7 & 63.0 & 76.6  & 87.3  & 90.8   \\
    MEB-Net w/o \textit{AR}  & 75.5 & 89.3 & 95.9 & 97.4  & 65.4 & 77.9 & 88.9 & 91.9  \\
    \hline  
    MEB-Net     & 76.0 & 89.9 & 96.0 & 97.5  & 66.1 & 79.6 & 88.3 & 92.2  \\
    \hline\hline
\end{tabular}
\end{center}

\caption{Ablation studies: \textit{Supervised Models:} - Re-ID models trained using the labelled target-domain training images. \textit{Direct Transfer:} - Re-ID models trained by labelled source-domain training images. $\mathcal{L}_{vot}$ (Eq.~\ref{eq:vot}), $\Theta_T$ (Eq.~\ref{eq:average}), $\mathcal{L}_{mid}$ (Eq.~\ref{eq:mid}) and  $\mathcal{L}_{mtri}$ (Eq.~\ref{eq:mtri}) are described in Sec. \ref{sec:bs}. \textit{AR:} Authority Regularization as described in Sec.~\ref{sec:ar}.}
\label{table:ablation}

\end{table*}

Detailed ablation studies are performed to evaluate the components of MEB-Net as shown in Table~\ref{table:ablation}.

\textbf{Supervised models vs. Direct transfer.} 
We first derive the upper and lower performance bounds by the supervised models (trained using labelled target-domain images) and the direct transfer models (trained using labelled source-domain images) for the ablation studies as shown in Table~\ref{table:ablation}. 
We evaluate all three architectures and report the best results in Table~\ref{table:ablation}. 
It can be observed that the huge performance gaps between the Direct Transfer models and the Supervised Models due to the domain shift. 

\textbf{Voting loss:} 
We create baseline ensemble models that only use voting loss. 
Specifically, pseudo-labels are predicted by averaging the features outputted from all expert networks, and then used to supervise the training of each expert network individually by optimizing the voting loss. 
As Table \ref{table:staart} shows, the Baseline model outperforms the Direct Transfer model by a large margin. 
This shows that the voting loss effectively make use of the ensemble models to predict more accurate pseudo-labels and fine-tune each network. 

\textbf{Temporally Average Networks:}
The model removing the temporally average models is denoted as "MEB-Net w/o $\Theta_T$". For this experiment, we directly use the prediction of the current networks parameterized by $\theta_T$ instead of the temporally average networks with parameters $\Theta_T$ as soft labels. As Table. \ref{table:ablation} shows, distinct drops of 5.3\% mAP and 2.8\% rank-1 accuracy are observed for Market1501$\rightarrow$DukeMTMC-reID transfer.
Without using temporally average models, networks tend to degenerate to be homogeneous, which substantially decreases the learning capability.

\textbf{Effectiveness of mutual learning:} 
We evaluate the mutual learning component in Sec. \ref{sec:bs} from two aspects: the mutual identity loss and the mutual triplet loss. The former is denoted as "MEB-Net w/o $\mathcal{L}_{mid}$". Results show that mAP drops from 76.0\% to 70.2\% on Market-1501 dataset and from 66.1\% to 60.4\% on DukeMTMC-reID dataset. Similar drops can also be observed when studying the mutual triplet loss, which are denoted as "MEB-Net w/o $\mathcal{L}_{mtri}$". For example, the mAP drops to 74.9\% and 63.0\% for DukeMTMC-reID$\rightarrow$Market-1501 and vice versa, respectively. The effectiveness of the mutual learning, including both two mutual loss, can be largely attributed to that it enhances the discrimination capability of all expert networks.

\textbf{Authority Regularization:}
We verify the effectiveness of the proposed authority regularization (Sec~\ref{sec:ar}) of MEB-Net. Specifically, we remove the authority regularization, and set authority $w=1$ (in Eq.~\ref{eq:arid} and Eq.~\ref{eq:artri}) equally for all expert models. The model is denoted as "MEB-Net w/o \textit{AR}", of which the results are shown in Table \ref{table:staart}.
Experiments without authority regularization shows distinct drops on both Market-1501 and DukeMTMC-reID datasets, which indicates that equivalent brainstorming among experts hinders feature discrimination because an unprofessional expert may provide erroneous supervision. 

\subsection{Discussion}

\textbf{Comparison with Baseline Ensemble.}
\begin{table*}[t]
\begin{center}
\begin{tabular}{l|p{1.8cm}<{\centering}|p{1.8cm}<{\centering}|p{1.8cm}<{\centering}|p{1.8cm}<{\centering}|p{1.8cm}<{\centering}}
    \hline\hline
    Architectures & Supervised & Dire. tran. & Sing. tran. & Base. ens. & MEB-Net \\ 
    \hline\hline
    DenseNet-121    & \textbf{80.0}  & 30.8 & 57.8 & 69.5 & \underline{76.0} \\
    \hline
    ResNet-50       & \textbf{82.5}  & 31.5 & 62.4 & 65.6 & \underline{72.2}\\
    \hline
    Inception-v3    & \textbf{68.3}  & 28.5 & 51.5 & 62.3 & \underline{71.3}\\
\hline\hline
\end{tabular}
\end{center}

\caption{mAP ($\%$) of networks of different architectures for DukeMTMC-reID $\rightarrow$ Market-1501 transfer: \textit{Supervised} - supervised models; \textit{Dire. tran.} - direct transfer; \textit{Sing. tran.} - single model transfer; \textit{Base. ens.} - baseline ensemble.}
\label{table:arens} 

\end{table*} 

Considering that ensemble models usually achieve more superior performance than a single model, we compare mAPs of our approach with other baseline methods, including single model transfer and baseline model ensemble.
Results are shown in Table.~\ref{table:arens}. 
%
%
%
The baseline model ensemble uses all networks to extract average features of unlabelled samples for pseudo-label prediction, but without mutual learning among them while adaptation in the target domain. 
The improvement of baseline ensemble than single model transfer is because of more accurate pseudo-labels. However, MEB-Net performs significantly better than all compared methods. 
It validates that MEB-Net provides a more effective ensemble method with respect to domain adaptive person re-ID.

\begin{figure}[t]
\centering
\includegraphics[width=0.9\linewidth]{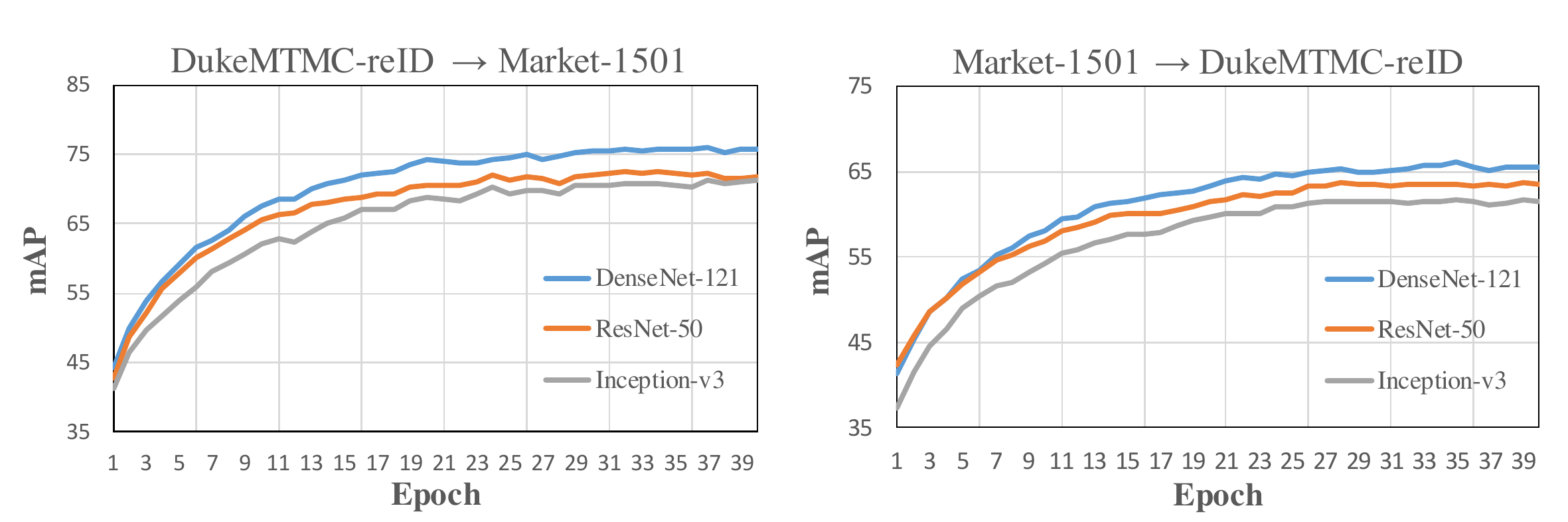}

\caption{Evaluation with different \textit{epoch}. The performance of all networks ascend to a stable value after 20 epochs.}
\label{fig:epoch}

\end{figure}

\noindent\textbf{Number of Epochs.} We evaluate the mAP of MEB-Net after each epoch, respectively. As shown in Fig. \ref{fig:epoch}, the models become stronger when the iterative clustering proceeds. The performance is improved in early epochs, and finally converges after 20 epochs for both datasets.

\section{Conclusion}

The paper proposed a multiple expert brainstorming network (MEB-Net) for domain adaptive person re-ID. MEB-Net adopts a mutual learning strategy, where networks of each architecture are pre-trained to initialize several expert models while the adaptation is accomplished through brainstorming (mutual learning) among expert models. Furthermore, an authority regularization scheme was introduced to tackle the heterogeneity of experts. Experiments demonstrated the effectiveness of MEB-Net for improving the discrimination ability of re-ID models. Our approach efficiently assembled discrimination capability of multiple networks while requiring solely a single model during inference time throughout. 

\section*{Acknowledgement}

This work is partially supported by grants from the National Key R\&D Program of China under grant 2017YFB1002400, the National Natural Science Foundation of China (NSFC) under contract No. 61825101, U1611461 and 61836012.

%
%
\bibliographystyle{splncs04}
\bibliography{egbib}
\end{document}